\documentclass[preprint,12pt]{elsarticle}

\usepackage{amssymb}
\usepackage{amsmath}
\usepackage[justification=centering]{caption} 
\usepackage{caption} 
\usepackage{array} 
\usepackage{booktabs}
\usepackage{multirow}
\usepackage{graphicx}   
\usepackage{booktabs}   
\usepackage{makecell}   
\usepackage{xcolor}     
\usepackage{tabularx} 

\journal{Nuclear Physics B}

\begin{document}

\begin{frontmatter}



\title{Sample-Centric Multi-Task Learning for Detection and Segmentation of Industrial Surface Defects}


\author[label1,label2]{Hang-Cheng Dong}
\author[label3]{Yibo Jiao}
\author[label3]{Fupeng Wei}
\author[label1,label2]{Guodong Liu}
\author[label1]{Dong Ye}
\author[label1,label2]{Bingguo Liu\corref{cor2}}
\ead{liu_bingguo@hit.edu.cn}
\cortext[cor2]{Corresponding Author}
\affiliation[label1]{organization={School of Instrumentation Science and Engineering, Harbin Institute of Technology},
            addressline={Xidazhi 92}, 
            city={Harbin},
            postcode={150001}, 
            state={Heilongjiang},
            country={China}}
            
\affiliation[label2]{organization={Harbin Institute of Technology Suzhou Research Institute},
            addressline={Nanguandu 500}, 
            city={Suzhou},
            postcode={215100}, 
            state={Jiangsu},
            country={China}}

\affiliation[label3]{organization={School of Information Engineering, North China University of Water Resources and Electric Power},
            addressline={No.136, East Jinshui Road}, 
            city={Zhengzhou},
            postcode={450046}, 
            state={Henan},
            country={China}}

\begin{abstract}
	Industrial surface defect inspection for sample-wise quality control (QC) must simultaneously decide whether a given sample contains defects and localize those defects spatially. In real production lines, extreme foreground–background imbalance, defect sparsity with a long-tailed scale distribution, and low contrast are common. As a result, pixel-centric training and evaluation are easily dominated by large homogeneous regions, making it difficult to drive models to attend to small or low-contrast defects—one of the main bottlenecks for deployment. Empirically, existing models achieve strong pixel-overlap metrics (e.g., mIoU) but exhibit insufficient stability at the sample level, especially for sparse or slender defects. The root cause is a mismatch between the optimization objective and the granularity of QC decisions. To address this, we propose a sample-centric multi-task learning framework and evaluation suite. Built on a shared-encoder architecture, the method jointly learns sample-level defect classification and pixel-level mask localization. Sample-level supervision modulates the feature distribution and, at the gradient level, continually boosts recall for small and low-contrast defects, while the segmentation branch preserves boundary and shape details to enhance per-sample decision stability and reduce misses. For evaluation, we propose decision-linked metrics, Seg\_mIoU and Seg\_Recall, which remove the bias of classical mIoU caused by empty or true-negative samples and tightly couple localization quality with sample-level decisions. Experiments on two benchmark datasets demonstrate that our approach substantially improves the reliability of sample-level decisions and the completeness of defect localization.
\end{abstract}


\end{frontmatter}



\section{Introduction}
\label{sec1}

Industrial visual inspection is a cornerstone of smart manufacturing, trending toward online operation, full inspection, and automated decision-making\cite{prunella2023deep}. However, surface defects (e.g., cracks, scratches, point-like blemishes) often exhibit extreme foreground–background imbalance, long-tailed shape/scale distributions, and low visual contrast. These statistics cause pixel-centric training and evaluation to be dominated by large normal regions, making it difficult to capture small or faint defects and thereby increasing the cost of false negatives in production\cite{tabernik2020segmentation}. Consequently, improving sample-level decision reliability while preserving pixel-level localization has become the central challenge for bringing industrial defect detection to engineering deployment.

Existing industrial defect detection algorithms largely follow a coarse-to-fine paradigm. Image classification provides sample-level “defect/no-defect” decisions with high efficiency but no spatial localization\cite{yang2019real}. Object detection further offers instance-level localization with coarse boundaries, making it suitable for rapid triage of discrete or sparse defects. In contrast, semantic segmentation outputs pixel-accurate defect morphology, supporting dimensional quantification, area statistics, and rework planning, and is increasingly becoming a key module for precise contouring and process traceability. Additionally, anomaly detection constitutes an alternative, which is appropriate when defect categories are difficult to predefine. Among these approaches, semantic segmentation is central in industrial scenarios that require precise defect boundaries due to its highest spatial resolution and directly quantifiable mask outputs\cite{wang2018fast,luo2020automated}.

By annotation reliance and learning paradigm, existing segmentation methods can be grouped into fully supervised and weakly supervised approaches. Fully supervised methods depend on finely annotated pixel masks and typically adopt encoder–decoder architectures such as FCN\cite{long2015fully}, U-Net\cite{ronneberger2015u}, or DeepLab\cite{chen2017rethinking}. Under joint cross-entropy and Dice-style losses, they learn detailed boundaries and morphological features. The resulting masks are high quality and geometrically faithful, facilitating downstream quantification and process control; however, they entail high labeling cost and are susceptible to label noise introduced by annotator subjectivity\cite{brust2018active}.

In annotation-constrained scenarios, weakly supervised segmentation offers a complementary path that targets an engineering trade-off between labeling cost and accuracy. These methods use weak signals—e.g., image-level tags, bounding boxes, or point annotations—as priors, and iteratively refine predicted boundaries via techniques such as pseudo-label self-training and consistency regularization\cite{wei2016stc,jiao2025fgr}; alternatively, a small amount of pixel-level supervision can guide learning over large unlabeled sets\cite{papandreou2015weakly,wei2025weakly}. This paradigm fits production data characteristics—abundant normal samples, few defective cases, and scarce fine annotations—but its mask quality typically trails fully supervised methods in boundary precision and topological completeness, and it is more sensitive to prior design and post-processing thresholds\cite{gozzovelli2021tip}.

Although learning strategies vary, evaluation for industrial defect segmentation still largely borrows the mean Intersection-over-Union (mIoU) used in natural-image tasks\cite{everingham2010pascal,cordts2016cityscapes,long2015fully}. While general and easy to implement, mIoU is semantically misaligned with industrial defect detection and quality inspection: First, extreme foreground–background imbalance and a large fraction of empty samples cause true negatives to dominate, artificially inflating aggregate scores—so much so that a complete miss can still yield a deceptively high mIoU\cite{eelbode,taha2015metrics}. Second, the continuous average lacks a direct linkage to production pass/fail thresholds and first-pass yield (FPY), whereas real manufacturing relies on thresholded, sample-level decisions—mIoU provides weak support for such process control\cite{fawcett2006introduction,bergmann2019mvtec}.

To overcome the systematic bias and semantic misalignment of mIoU in industrial segmentation evaluation, we propose a practical, industry-oriented metric—Sample-level mean Intersection-over-Union (Sampl\_mIoU). Using the image as the statistical unit, the Sample\_mIoU macro-averages the foreground IoU over only defect-containing samples, while reporting defect-free samples as a separate category. This “positive-only scoring with a separate track for empty samples” avoids dilution and artificial inflation from numerous true negatives, directly penalizes misses at the sample level, and improves robustness to shifts in dataset composition. Moreover, distributions derived from Sample\_mIoU—e.g., Seg\_Accuracy—map naturally to shop-floor KPIs such as conformance rate and FPY, thereby achieving interpretable alignment with production decision semantics.
Moreover, although Sample\_mIoU effectively remedies the above evaluation shortcoming, improving the metric alone is not a substitute for model-level optimization. Under the conventional training paradigm of a single segmentation or classification network, a pixel-centric loss causes gradients to be dominated by large target regions; responses to minute defects are systematically suppressed, which ultimately appears as recall shortfalls on Sample\_mIoU.

To address this, we propose a sample-centric multi-task learning (MTL) framework: a sample branch and a segmentation branch produce parallel outputs and are jointly optimized. End-to-end training adopts a composite loss that explicitly couples the sample-level global decision—whether a defect exists—with the pixel-level fine-grained delineation—where the defect is. This design allows sample-level supervision, via the shared encoder, to impose backpropagated constraints and informational guidance on the segmentation branch, maintaining high classification accuracy while improving segmentation completeness and the detection rate of small defects. Our core contributions are as follows:

\begin{enumerate}
	\item This work proposes the Sample\_mIoU evaluation metric, which treats each image as the statistical unit and macro-averages the foreground IoU over positive samples, eliminating score dilution and artificial inflation caused by foreground–background pixel imbalance and empty images, thereby explicitly penalizing misses at the sample level.
	
	\item Building on Sample\_mIoU, we further define Seg\_Accuracy and its accompanying specification, enabling evaluation results to map directly to the pass rate used on production lines, thereby improving alignment and interpretability for shop-floor decisions.
	
	\item We propose a sample-centric, plugin-based MTL scheme: a lightweight sample-level head attached to the shared encoder is jointly optimized with the segmentation branch. By injecting sample-level supervision aligned with shop-floor decisions, it mitigates gradient dominance caused by foreground–background imbalance and—without sacrificing classification accuracy—improves segmentation completeness and small-defect recall, with virtually no additional parameters or compute overhead.
\end{enumerate}

\section{Related Work}\label{sec2}

\subsection{Classical Semantic Segmentation Algorithms}

Industrial surface defect inspection simultaneously requires a reliable sample-level decision on defect presence and precise pixel-level localization of the defect extent. In real production settings, however, data typically exhibit a high background ratio, sparse defects with long-tailed scales, low contrast, and condition/domain drift. Under this regime, relying solely on pixel-weighted losses and metrics fails to adequately expose the miss-detection (false-negative) risk of small yet critical defects, creating a mismatch between the granularity of evaluation and per-item quality control. To address this, research on semantic segmentation has evolved along two complementary lines—fully supervised and weakly supervised—balancing representation quality against annotation cost.

Fully supervised approaches utilize encoder-decoder architectures as their foundation, effectively improving spatial consistency and boundary compliance through multi-scale context and high-resolution details. The U-Net\cite{ronneberger2015u} architecture effectively integrates global semantics with local texture via symmetric convolutional layers with skip connections, hence serving as a foundational benchmark for small-sample and fine-grained segmentation. SegNet\cite{badrinarayanan2017segnet} utilizes pooling indices to direct upsampling, reducing reconstruction blur while enhancing memory efficiency. PSPNet\cite{chen2018encoder} utilizes pyramid pooling to consolidate global context and address single-scale semantic limitations, whereas DeepLab v3+\cite{zhao2017pyramid} integrates a dilated spatial pyramid with a streamlined decoder to broaden the receptive field and improve boundary refinement. HRNet\cite{wang2020deep} utilizes parallel multi-resolution branches with ongoing information sharing to preserve strong high-resolution representations, enhancing its capability to address complicated boundaries and intricate structures. Although these methodologies markedly improve pixel-level overlap, their objective functions and assessment metrics continue to focus on pixel consistency. As a result, the risk of sample-level false negatives in situations with sparse faults or poor contrast is sometimes inadequately emphasized.

Weakly supervised semantic segmentation (WSSS) aims to reduce the cost of pixel-level annotation, typically using image-level labels to trigger class activation maps (CAMs) and then performing region expansion, affinity propagation, and boundary constraints to synthesize pseudo-masks. AffinityNet\cite{ahn2018learning} learns pixel-level semantic affinities and, together with a random-walk procedure, expands sparse class-activation seeds into more complete objects, while IRNet\cite{ahn2019weakly} further models inter-pixel relations and boundary cues explicitly to improve instance-level contours. Moreover, CAM, as a foundational mechanism for discriminative localization, has become a standard starting point in WSSS pipelines\cite{LUO2021107858}. Despite substantial progress on natural-scene segmentation, pseudo-label noise and coarse boundary delineation are amplified in industrial small-defect settings; training and evaluation remain centered on pixel aggregation, which makes it difficult to align with per-item decisions where the cost of missed detections is higher.

In summary, fully supervised methods are more stable in pixel-level overlap and boundary adherence, while weakly supervised methods offer advantages in annotation cost. A key challenge shared by both is that optimization objectives and evaluation criteria remain dominated by pixel-level granularity, making it difficult to explicitly constrain sample-level misses and keep alignment with quality-control decisions. This gap motivates the approach emphasized in this paper: introduce sample-level supervision at both the architectural and loss-design levels to exert persistent guidance on segmentation, and adopt sample-level metrics at evaluation that are consistent with per-item quality control—thereby achieving method–metric alignment across both detection and localization.

\subsection{Fine-Grained Augmentation for Semantic Segmentation}
Focusing on the detectability of small-target defect structures, existing research on the pixel side mainly strengthens representation and optimization along three directions: first, multi-scale and high-resolution representations preserve detail and improve boundary consistency by maintaining continuous information exchange across parallel resolutions, with HRNet as a representative that sustains high-resolution semantics via multi-branch collaboration \cite{wang2020deep}; second, attention and saliency mechanisms applied in skip connections or during decoding suppress background and highlight target regions—for example, Attention U-Net uses gated attention to filter cross-level features, and UNet++ employs nested dense skip connections to narrow the semantic gap and refine local structures \cite{oktay2018attention,zhou2018unet++}; third, loss-level treatments for class imbalance and boundary modeling replace pure region-overlap objectives with contour- or distance-based measures to handle extreme imbalance, where Boundary Loss optimizes a distance-to-curve function near the contour to improve boundary adherence and convergence stability \cite{kervadec2019boundary}. While these directions markedly improve pixel-level overlap and boundary quality, their supervisory signal remains centered on overlap fidelity, making it difficult to directly encode the sample-level risk constraint of no missed detection per workpiece.

To inject global semantics and decision cues at the representation level, MTL networks place related tasks—such as segmentation and classification—on top of a shared encoder and realize complementarity via joint optimization. Extensive evidence in medical imaging and industrial defect settings indicates that segmentation–classification synergy improves robustness and generalization; a common design uses globally pooled high-level semantics to assist segmentation, and then feeds segmentation priors back to classification \cite{he2023joint,zhou2021multi,graham2023one}. On the optimization side, strategies such as uncertainty weighting and gradient normalization dynamically balance branch gradients and suppress negative transfer \cite{graham2023one,chen2018gradnorm}. However, most works still adopt pixel-wise metrics or image-level ROC as the primary objective and evaluation axis; the sample-level miss-detection cost is not explicitly incorporated into the training dynamics, leaving a mismatch between metric and objective granularity under a per-item quality-control risk structure.

Building on this, we treat pixel-side fine-grained enhancements as a complementary, drop-in representational base, and embed sample-level supervision into the shared encoder to impose a global guidance signal. Compared with enhancement schemes that pursue pixel overlap alone or generic MTL frameworks that do not explicitly encode sample-level risk, our design surfaces the detection–localization joint objective as a first-class concern at both the architectural and loss-design levels, and closes the loop with sample-level metrics—thereby mechanistically mitigating the dilution effect of pixel pooling on small-scale, low-contrast defects.

\section{Methodology}

\subsection{Sample\_mIoU: How does the pixel level loss function affect the accuracy of defect detection}

\begin{figure*}[ht]
	\centering
	\includegraphics[width=0.6\textwidth, height=0.5\textheight]{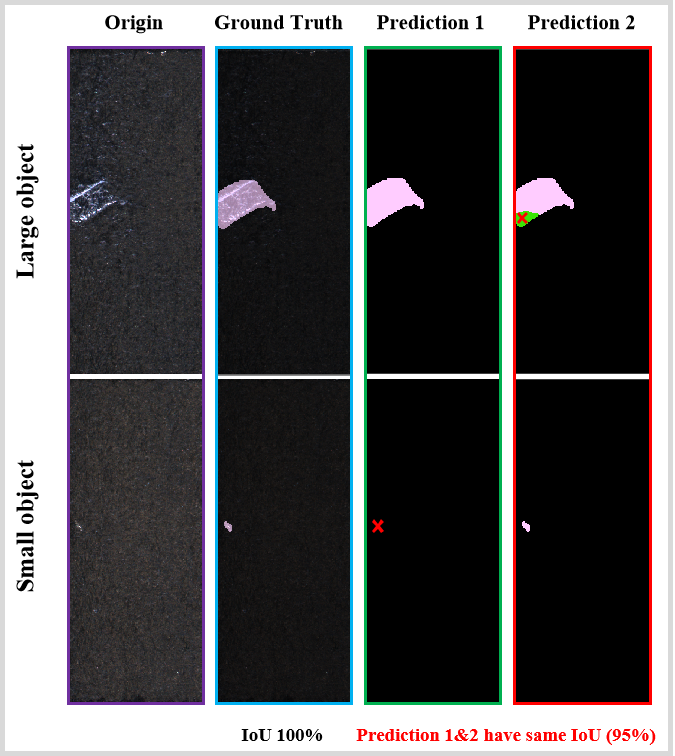} 
	\captionsetup{justification=centering}
	\caption{High Scores $\neq$ Good Models: The Disconnect Between Evaluation Metrics and Real-World Needs}
	\label{Figure.1.1} 
\end{figure*}

In the production-line setting of industrial surface defect inspection, the basic granularity of quality control and disposition decisions is the workpiece (i.e., the sample), rather than the pixel. A single miss immediately causes a sample-level false pass—and thus a quality risk—whereas false alarms typically incur reinspection or rework costs, yielding a markedly asymmetric cost structure. At the same time, images commonly exhibit an extremely high background ratio, sparse defects with a long-tailed scale distribution, and perturbations from glare, texture, and illumination drift. Under these statistics, any pixel-weighted metric is easily dominated by large background regions or large defects: as shown in Figure \ref{Figure.1.1}, even though the second image in Prediction1 fails to detect any defect, the first image contributes the vast majority of positive (defect) pixels, so the dataset-level TP/FP/FN tallies are governed almost entirely by it and the mIoU can still be as high as 95\%. This creates a granularity mismatch between evaluation and the sample-level acceptance used on the line, both masking the miss risk for small but critical defects and systematically underestimating the model’s detection capability on small defects.

To enhance accuracy in our discourse, we will initially delineate the segmentation overlap for each sample $i$. The intersection-over-union ratio in a sample can be articulated as:

\begin{equation}
		IoU_i=\frac{TP_i}{TP_i+FP_i+FN_i}
\end{equation}

In this context, $TP_i$ represents the "quantity of defect pixels accurately identified" in the sample, $FP_i$ signifies the "quantity of background pixels erroneously classified as defects," and $FN_i$ indicates the "quantity of true defect pixels overlooked." The denominator $TP_i + FP_i + FN_i$ represents the union of pixels identified by the model and those classified as true faults in the sample. Through comprehensive pixel-level measurement across all data, we derive the widely utilized (pixel pooling) mIoU:

\begin{equation}
	mIoU_{pooled}=\frac{\sum_i{TP_i}}{\sum_i{\left( TP_i+FP_i+FN_i \right)}}
\end{equation}

The numerator of this statistic represents the total number of pixels where the prediction coincides with the ground truth across all data, whilst the denominator signifies the total number of pixels included by both the prediction and ground truth across all data. This method, from a weighted average standpoint, is analogous to applying weights to $IoU_i$ based on the union pixel count of each sample. This pixel-based weighting inherently prioritizes "large target or large area" samples, potentially resulting in a "pixel dilution effect": even if multiple "small yet critical" flaws are entirely overlooked, the overall score may remain elevated as long as a few large defect samples are accurately identified.

To align the evaluation granularity with sample-level quality control, a straightforward approach is to first compute the IoU within each sample and then take an equal-weight average across samples (i.e., each workpiece is equally important):

\begin{equation}
	Sample\_mIoU_{hat}=\frac{1}{M}\sum_{i=1}^M{IoU_i}
\end{equation}

Where $M$ is the total number of samples. This form effectively weakens the pixel bias introduced by samples with large objects/areas, but two issues arise in derivation and implementation: first, when a sample contains neither ground-truth defects nor predicted defects, the denominator of $IoU_i$ is zero and the quantity is undefined (a pure true-negative sample); second, even if one assigns an ad-hoc score to such samples (e.g., 1 to denote “correctly predicting no defect,” or 0 to avoid artificial inflation), the resulting average loses comparability as the proportion of pure true negatives fluctuates, making the evaluation overly sensitive to dataset composition across batches or operating conditions.

To eradicate the previously mentioned uncertainty and sample composition bias, we ultimately employed a rigorous definition that averages solely “defect-related” samples: a sample is incorporated in the average only if it contains an authentic flaw or the model forecasts a defect. Denote the quantity of such valid samples as $M_{eff}$. The resilient sample-level average measure is subsequently defined as:

\begin{equation}
	Sample\_mIoU=\frac{1}{M_{eff}}\sum_{i\in relevant}{IoU_i}
\end{equation}

Here, “$relevant$” signifies samples in which the union of pixels exhibiting actual flaws and those projected to have defects is non-zero; $M_{eff}$ represents the quantity of such samples. After this modification, all $IoU_i$ values utilized in the computation are clearly specified, and the metric is no longer affected by the ratio of genuine negative samples. If any defective sample that should have been identified is overlooked (i.e., $IoU_i$ = 0), this outcome will equally affect the entire measure, directly indicating risks in sample-level quality control. To ensure statistical transparency, the ratio of true negatives may be reported independently alongside the metrics for process monitoring and regression analysis. If business requirements demand the representation of varying significance among samples or essential processes, non-zero weights may be incorporated into the numerator and denominator of the calculation formula to assign priority to pertinent samples while maintaining resilience against outliers.

\subsection{Seg\_Accuracy and Seg\_Recall}

In the production-line context of defect inspection, beyond localizing the spatial extent of defects, the more critical concern is the reliability of the sample-level defect/no-defect decision. This decision can be derived directly from the segmentation output, yielding sample-level classification metrics—collectively denoted Seg\_* (e.g., Seg\_Accuracy, Seg\_Recall).

Define the ground truth label of the $i$th sample as $y_i \in \{0,1\}$ (where 1 signifies the existence of a defect), and denote the model's output binary segmentation mask as $S_i$. The criterion for associating the segmentation mask with the sample-level prediction label $\hat{y}_i$ is:

\begin{equation}
	\hat{y}_i = \mathbf{I}\!\left\{\Phi(S_i) \ge \tau\right\}
\end{equation}

Here, $\Phi(\cdot)$ signifies the statistical function utilized for partition outcomes, whereas $\tau$ specifies the decision threshold. Using $\hat{y}_i$ and $y_i$, the quantities of $TP$, $TN$, $FP$, and $FN$ can be statistically calculated at the sample level, defined as follows:

\begin{equation}
	Seg\_Accuracy=\frac{TP+TN}{TP+TN+FP+FN}
\end{equation}

\begin{equation}
	Seg\_Recall=\frac{TP}{TP+FN}
\end{equation}

Here, Seg\_Accuracy characterizes the overall correctness of the sample-level decision, while Seg\_Recall emphasizes positive-class coverage, reflecting the model’s sensitivity to defective samples—i.e., its ability to suppress miss risk. Together with Sample\_mIoU, they provide complementary evaluation dimensions: the former answers “was the decision correct”, and the latter answers “given a positive sample, is the localization adequate”, quantified as the spatial overlap between the predicted mask and the ground truth on positives. In industrial scenarios where defects are sparse and the cost structure is asymmetric (misses cost more than false alarms), jointly reporting Seg\_Accuracy, Seg\_Recall, and Sample\_mIoU constrains both detection and localization, reduces metric bias from pixel pooling, aligns evaluation with per-item quality-control objectives, and supports the engineering goal of improving segmentation quality under reliable sample-level decisions.

\begin{figure*}[ht]
	\centering
	\includegraphics[width=1\textwidth, height=0.33\textheight]{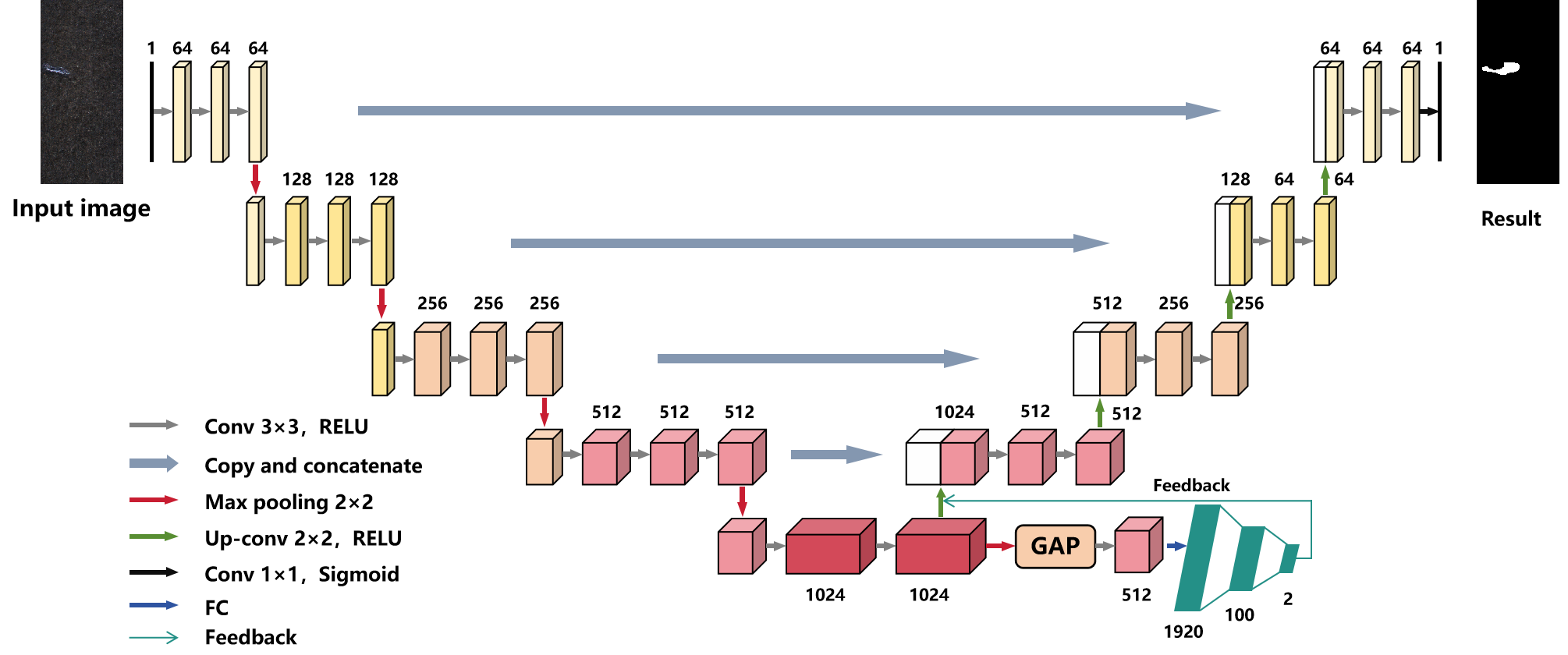} 
	\captionsetup{justification=centering}
	\caption{MTL Network: Joint Segmentation–Classification Learning on a Shared Encoder}
	\label{Figure.1} 
\end{figure*}

\subsection{MTL Network}

In industrial defect segmentation, when training and evaluation are pixel-centric, gradients and parameter updates are dominated by large regions, which systematically dilutes sensitivity to small/low-contrast defects—an effect we term pixel bias. A direct consequence is that even with a strong overall mIoU, any sample whose critical region is missed will have near-zero IoU, and the macro average across samples is markedly reduced.

To mitigate this mismatch induced by data statistics and pixel weighting, we propose a sample-centric loss implemented as a sample-aware multi-task branch on the shared encoder (as depicted in Figure \ref{Figure.1}). The segmentation branch provides dense spatial supervision, while the sample branch predicts defect presence/absence at the sample level and backpropagates through the shared encoder to steer representation learning—continually exerting recall pressure on “small but critical” targets. This improves Sample\_mIoU and sample-level recall without sacrificing pixel-wise overlap.

Built on a U-Net backbone, the encoder captures multi-scale context, and the decoder—via skip connections—restores spatial detail. Adding a classification head enables globally pooled signals to be derived from high-level features, the segmentation mask, or their fusion, producing a sample-level decision. Through joint optimization, the two supervision sources are aligned within a shared feature space: the segmentation branch is discouraged from optimizing solely for large-area overlap, while the classification branch is discouraged from relying only on global statistics and ignoring spatial consistency. Working in concert, they align the training objective with sample-level risk, with gains manifested as higher Sample\_mIoU and improved sample-level recall.

In the specified MTL architecture, segmentation and classification tasks utilize the same encoder and undergo joint optimization during training.
Denote the image domain of the $i$th sample as $\varOmega_i$, characterized by dimensions H×W pixels, so $|\varOmega_i|=HW$. Define the ground truth mask as $Y_i\in \left\{ 0,1 \right\} ^{H\times W}$, the defect probability map produced by the segmentation branch as $S_i\in \left[ 0,1 \right] ^{H\times W}$, the sample-level ground truth label as $y_i\in \left\{ 0,1 \right\}$, and the defect probability generated by the classification branch as $\hat{p}_i \in \left[ 0,1 \right]$. The composite loss is articulated as follows:

\begin{equation}
	L=\lambda _{seg}L_{seg}+\lambda _{cls}L_{cls},\ \lambda _{seg}=\lambda _{cls}=0.5
\end{equation}

The segmentation loss component utilizes pixel-level binary cross-entropy, averaged throughout the image domain $\varOmega_i$, to enforce spatial overlap and border consistency:

\begin{multline}
	L_{seg}=\frac{1}{HW}\sum_{x\in \varOmega_i}
	\Big[ -Y_i(x)\log S_i(x) - (1 - Y_i(x))\log\big(1 - S_i(x)\big) \Big]
\end{multline}

We adopt a sample-level binary cross-entropy as the classification loss to improve the reliability of the defect vs. non-defect decision:

\begin{equation}
	L_{cls}=-y_i\log \hat{p}-\left( 1-y_i \right) \log \left( 1-\hat{p}_i \right) 
\end{equation}

The composite objective aligns the sample-level detection signal with pixel-level spatial supervision within a shared feature space: $L_{seg}$preserves fine-grained structure and boundary information, suppressing the tendency to obtain high scores by focusing only on large, easy regions; $L_{cls}$ propagates the detection requirement backward into the encoder features and mask-generation process, imparting persistent gradients that target small and low-contrast defects. Their synergy ultimately manifests as improvements in Sample\_mIoU and sample-level recall.
\begin{figure*}[htbp]
	\centering
	\includegraphics[width=0.6\textwidth, height=0.22\textheight]{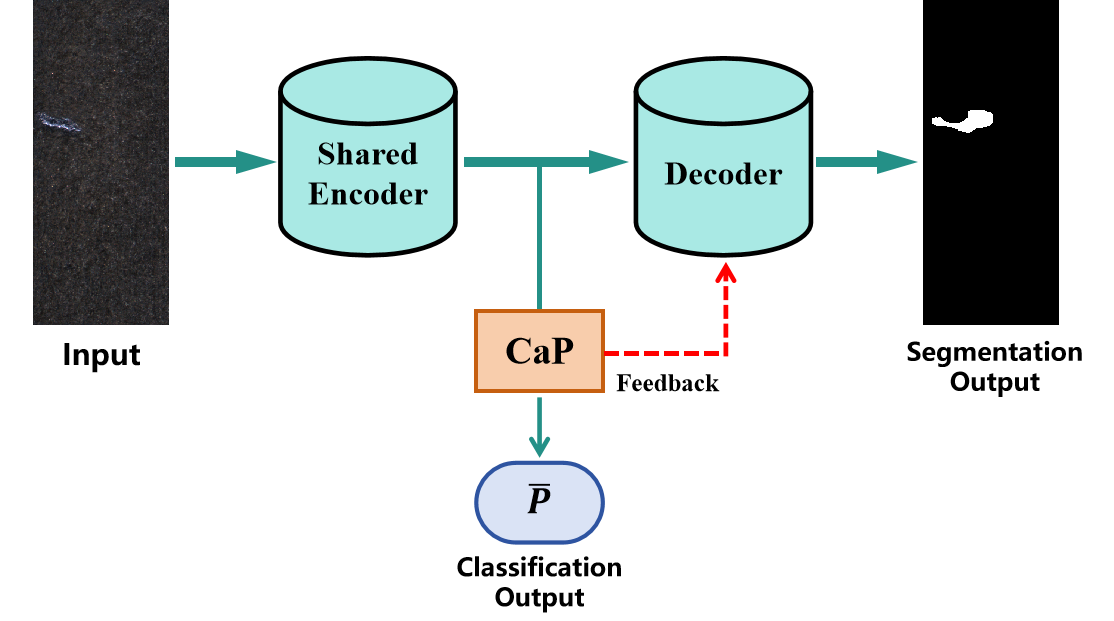} 
	\captionsetup{justification=centering}
	\caption{Abstract Structure and Information Flow of the CaP Plugin.}
	\label{CaP} 
\end{figure*}



\subsection{Classifier-as-Plugin Branch}
We abstract the new classifier as a generic, pluggable module—Classifier as a Plugin (CaP)—which injects the critical sample-centric supervision signal. As illustrated in Figure \ref{CaP}, a lightweight sample branch is attached to the output of the shared encoder of any segmentation backbone to produce a sample-level defect-presence decision, where $\bar{P}$ denotes the predicted probability of the class. Training uses a joint-loss objective in which gradients from the sample-level loss are backpropagated only into the encoder, leaving the decoder’s architecture and training procedure untouched. This allows the sample-side supervision to continuously steer the learned representations, mitigating the small/low-contrast defect sensitivity shortfall induced by pixel bias.

With a standardized interface, CaP integrates seamlessly into a range of segmentation architectures (e.g., U-Net, DeepLab, SegNet) with effectively negligible compute and parameter overhead. Ablation studies in Section \ref{sec5} show consistent gains in both sample-level detection and the localization of small/weak defects after adding CaP. The underlying mechanism is that the sample-centric loss and the pixel-wise segmentation loss co-optimize within the shared encoder, aligning the training objective with sample-level risk; this alignment ultimately manifests as improvements in Sample\_mIoU and sample-level recall. For deployment, we recommend attaching the sample branch to the topmost encoder stage or to multi-scale aggregated features. A default joint-loss weighting of 1:1 works well; under severe class imbalance, adapt the weights by the positive/negative sample ratio or a cost-sensitive scheme—and in recall-prioritized scenarios where missed detections are costly, modestly increase the sample-side weight.

\section{Experimental Data}

\subsection{Datasets}

This study employs three benchmark datasets—KolektorSDD2 \cite{bovzivc2021mixed}, and the Crack defect dataset—to train and evaluate the MTL model and metrics such as Sample\_mIoU and Seg\_Accuracy.

\textbf{KolektorSDD2} contains 3,335 high-resolution images (approximately 230 × 630 pixels), of which 356 are positive samples with surface defects (e.g., scratches, spots) and 2,979 are negative. The dataset is partitioned into a training set (246 positives / 2,085 negatives) and a test set (110 positives / 894 negatives). All defects are annotated with fine-grained, pixel-level segmentation masks, supplying precise supervisory signals.

\textbf{Crack} consists of 918 high-resolution road-defect images, including both cracked and non-cracked samples across diverse real-world road conditions. Images are provided at two resolutions—320 × 480 pixels and 256 × 256 pixels—and each image is annotated at the pixel level, accurately delineating the exact shape and location of crack regions.

    \subsection{Evaluation Metrics}

This work utilizes four metrics—mIoU, Sample\_mIoU, Seg\_Accuracy, and Seg\_Recall—to concurrently address pixel-level overlap quality and sample-level decision reliability, thus fulfilling the twin criteria of “sufficient localization” and “per-item judgment” in Industrial visual inspection contexts.
\begin{figure*}[htbp]
	\centering
	\includegraphics[width=0.7\textwidth, height=0.65\textheight]{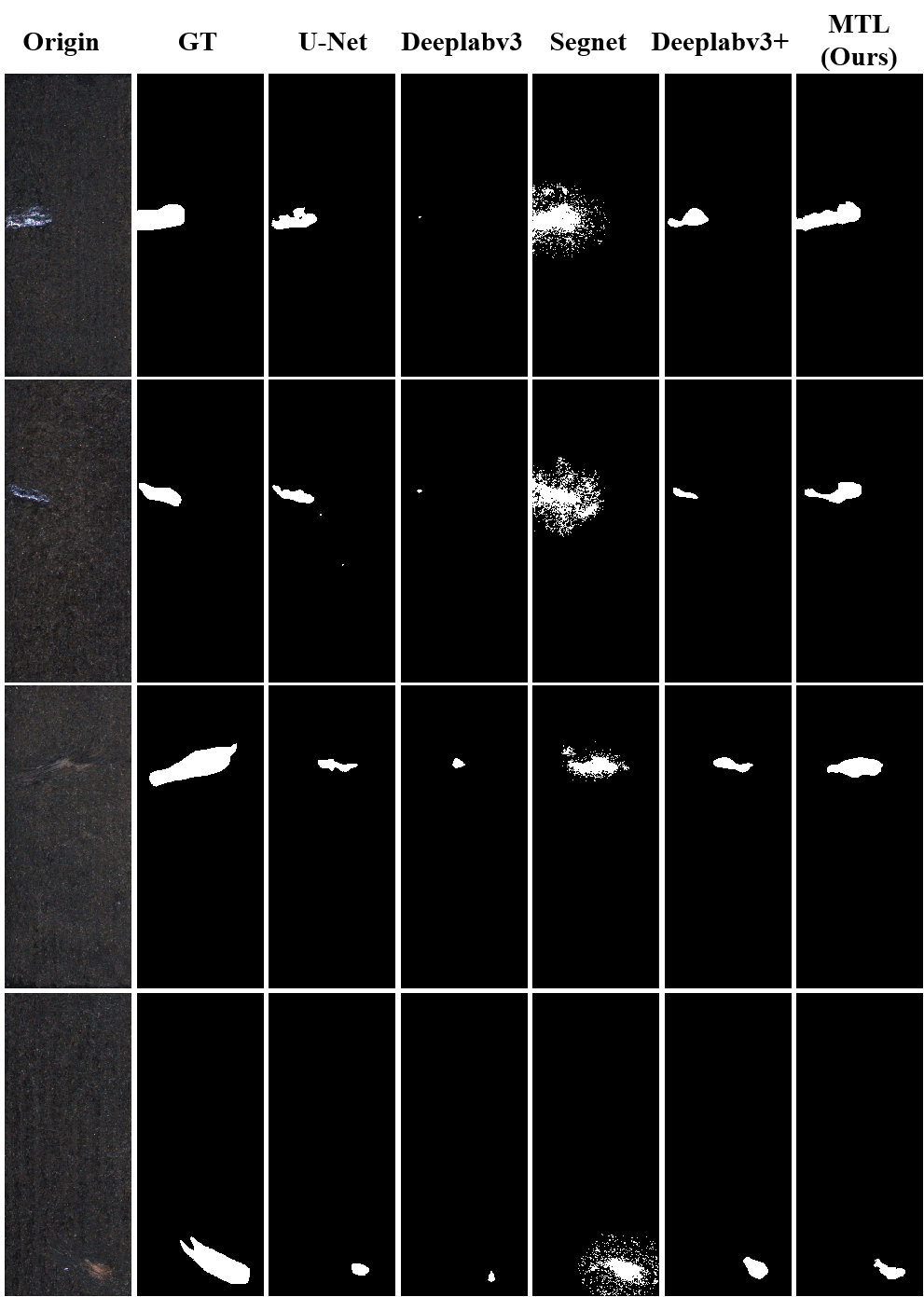} 
	\captionsetup{justification=centering}
	\caption{Segmentation Performance Comparison of Fully Supervised Models on the KSDD2 Dataset}
	\label{Figure.3} 
\end{figure*}

\begin{equation}
	mIoU_{pooled}=\frac{\sum_i{TP_i}}{\sum_i{\left( TP_i+FP_i+FN_i \right)}}
\end{equation}

\begin{equation}
	Sample\_mIoU=\frac{1}{M_{eff}}\sum_{i\in relevant}{IoU_i}
\end{equation}

\begin{equation}
	Seg\_Accuracy=\frac{TP+TN}{TP+TN+FP+FN}
\end{equation}

\begin{equation}
	Seg\_Recall=\frac{TP}{TP+FN}
\end{equation}

In this context, TP, FN, TN, and FP signify true positives, false negatives, true negatives, and false positives, respectively; $M_{eff}$ indicates the total number of effective samples inside the current category; $i$ represents the sample index. These four metrics enhance one another in evaluative dimensions: mIoU and Sample\_mIoU highlight the quality of spatial localization, whilst Seg\_Accuracy and Seg\_Recall define the overall accuracy of sample-level classification and the efficacy of positive class detection. Joint reporting supplies evidence regarding both the “localization” and “classification” aspects, delivering a more thorough assessment of the model's usefulness and resilience in actual production line settings.
\begin{table*}[!t]
  \centering
  \small
  \caption{Quantitative Performance Comparison (\%) of Different Defect Detection Models across Multiple Datasets}
  \begin{tabularx}{\textwidth}{>{\centering\arraybackslash}p{1.5cm}
                              >{\centering\arraybackslash}p{2.6cm}
                              *{4}{>{\centering\arraybackslash}X}}
    \toprule
    \textbf{Datasets} & \textbf{Models} & \textbf{mIoU} & \textbf{\makecell{Sample\\mIoU}} & \textbf{Seg\_Acc} & \textbf{Seg\_Recall} \\
    \midrule
    \multirow{5}{*}{KSDD2} 
      & Deeplab v3   & 52.38\% & 51.68\% & 89.34\% & 71.73\% \\
      & Deeplab v3+  & 53.24\% & 55.93\% & 92.03\% & 72.73\% \\
      & Segnet       & 55.05\% & 58.53\% & 90.04\% & 81.82\% \\
      & Unet         & 57.44\% & 60.37\% & 90.64\% & 90.00\% \\
      & \textbf{MTL (Ours)} & \textbf{61.74\%} & \textbf{62.08\%} & \textbf{92.93\%} & \textbf{85.45\%} \\
    \midrule
    \multirow{5}{*}{Crack} 
      & Deeplab v3   & 66.15\% & 77.53\% & 96.51\% & 100.00\% \\
      & Deeplab v3+  & 66.96\% & 73.25\% & 97.60\% & 99.68\% \\
      & Segnet       & 70.17\% & 77.93\% & 95.86\% & 99.84\% \\
      & Unet         & 69.71\% & 78.19\% & 97.82\% & 99.68\% \\
      & \textbf{MTL (Ours)} & \textbf{70.38\%} & \textbf{78.20\%} & \textbf{96.08\%} & \textbf{100.00\%} \\
    \bottomrule
  \end{tabularx}

  \label{tab:performance}
\end{table*}

\begin{figure*}[htbp]
	\centering
	\includegraphics[width=0.9\textwidth, height=0.4\textheight]{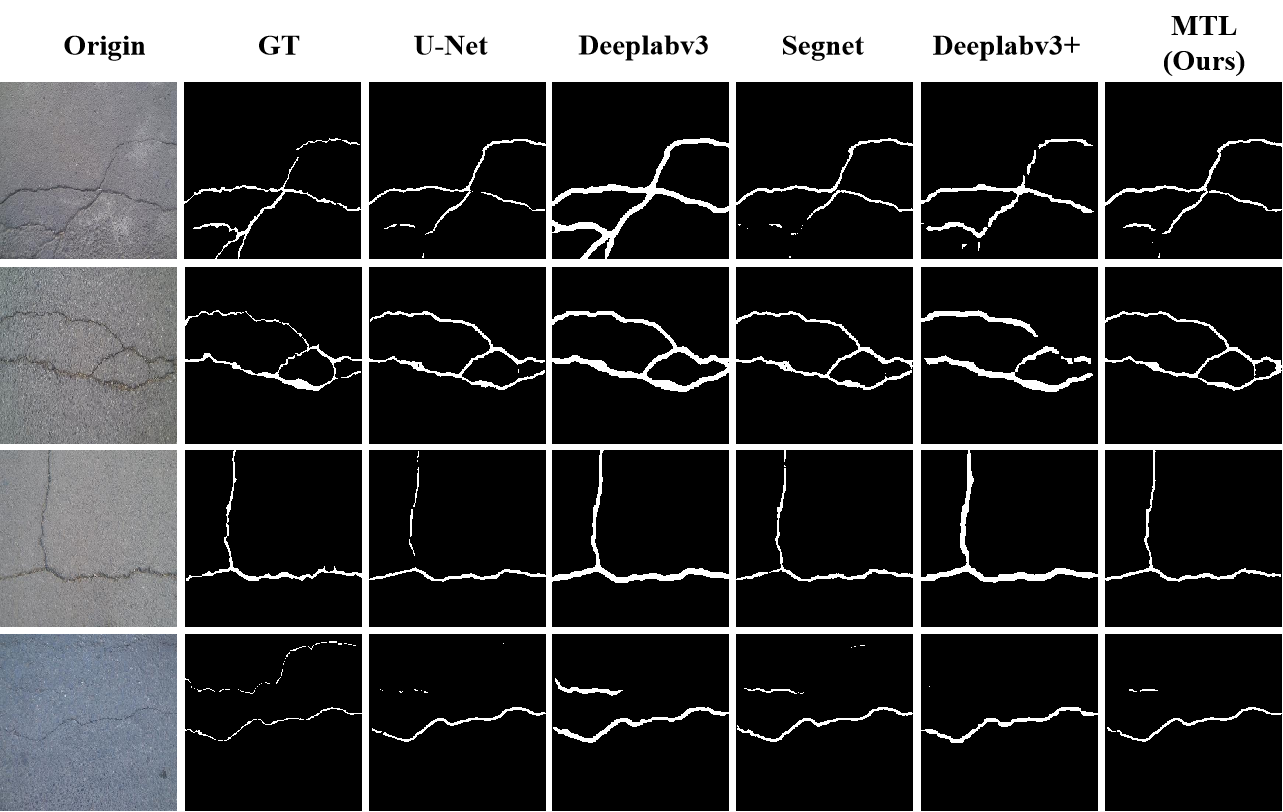} 
	\captionsetup{justification=centering}
	\caption{Segmentation Performance Comparison of Fully Supervised Models on the Crack Dataset}
	\label{Figure.4} 
\end{figure*}
\section{Experiment}\label{sec5}

We comparatively evaluate the proposed MTL network on two benchmark datasets against fully supervised baselines—including U-Net, SegNet, PSPNet, and DeepLabv3+ \cite{ronneberger2015u, chen2017rethinking,badrinarayanan2017segnet,zhao2017pyramid}. For controlled ablations, we append a lightweight classification branch to the encoder side of each baseline while keeping the decoder unchanged and the training protocol identical. Evaluation spans mIoU, Sample\_mIoU, Seg\_Accuracy, and Seg\_Recall, with an emphasis on whether sample-level classification cues, backpropagated into the encoder features and mask-generation process, improve sensitivity to small and low-contrast defects.

\begin{table*}[t]
\centering
\caption{\centering Ablation Results of Introducing the CaP Plugin Across Different Segmentation Architectures}
\label{tab:compare-ksdd2-crack}
\setlength{\tabcolsep}{4pt}
\renewcommand{\arraystretch}{1.15}
\makebox[\linewidth][c]{%
\resizebox{\linewidth}{!}{%
\begin{tabular}{c *{2}{cccc}}
\toprule
\multirow{2}{*}{\makecell[c]{Datasets \&\\ Models \& Metrics}} &
\multicolumn{4}{c}{\textbf{KSDD2}} &
\multicolumn{4}{c}{\textbf{Crack}} \\
\cmidrule(lr){2-5}\cmidrule(l){6-9}
& mIoU & Sample\_mIoU & Seg\_Acc & Seg\_Recall
& mIoU & Sample\_mIoU & Seg\_Acc & Seg\_Recall \\
\midrule
Deeplab v3
& 52.38\% & 51.68\% & 89.34\% & 71.73\%
& 66.15\% & 77.53\% & 96.51\% & 100.00\% \\
{\color{red}\textbf{Deeplab v3 (+ Classifier)}}
& {\color{red}\textbf{53.71\% $\uparrow$}} & {\color{red}\textbf{53.27\% $\uparrow$}} & {\color{red}\textbf{90.84\% $\uparrow$}} & {\color{red}\textbf{65.45\%}}
& {\color{red}\textbf{66.16\% $\uparrow$}} & {\color{red}\textbf{72.79\% }} & {\color{red}\textbf{96.73\% $\uparrow$}} & {\color{red}\textbf{100.00\%}} \\
Deeplab v3+
& 53.24\% & 55.93\% & 92.03\% & 72.73\%
& 66.96\% & 73.25\% & 97.60\% & 99.68\% \\
{\color{red}\textbf{Deeplab v3+ (+ Classifier)}}
& {\color{red}\textbf{58.21\% $\uparrow$}} & {\color{red}\textbf{58.82\% $\uparrow$}} & {\color{red}\textbf{92.43\% $\uparrow$}} & {\color{red}\textbf{82.73\% $\uparrow$}}
& {\color{red}\textbf{69.20\% $\uparrow$}} & {\color{red}\textbf{77.17\% $\uparrow$}} & {\color{red}\textbf{97.06\% $\uparrow$}} & {\color{red}\textbf{100.00\% $\uparrow$}} \\
Segnet
& 55.05\% & 58.53\% & 90.04\% & 81.82\%
& 70.17\% & 77.93\% & 95.86\% & 99.84\% \\
{\color{red}\textbf{Segnet (+ Classifier)}}
& {\color{red}\textbf{61.85\% $\uparrow$}} & {\color{red}\textbf{59.36\% $\uparrow$}} & {\color{red}\textbf{91.83\% $\uparrow$}} & {\color{red}\textbf{88.18\% $\uparrow$}}
& {\color{red}\textbf{70.28\% $\uparrow$}} & {\color{red}\textbf{77.70\% $\uparrow$}} & {\color{red}\textbf{95.10\%}} & {\color{red}\textbf{100.00\% $\uparrow$}} \\
\bottomrule
\end{tabular}
}
}
\end{table*}

\subsection{Semantic Segmentation Results}

While the traditional U-Net performs strongly in pixel-level overlap and spatial localization for image segmentation, its reliance on skip connections alone to recover detail often biases optimization toward large, easy regions and neglects critical small targets or fine-grained defects. To address this, we propose an MTL architecture that augments the standard U-Net encoder–decoder with an additional classification branch. By deciding, for each sample, whether a defect is present, this branch provides sample-level defect/no-defect supervision and, via the shared encoder, backpropagates its signals into the segmentation task—thereby guiding mask prediction and improving sensitivity to small defects.	

The experimental results in Table \ref{tab:performance} indicate that, it can be observed that in the KolektorSDD2 dataset—where defect samples are relatively scarce—the Sample\_mIoU metric is lower than the mIoU for some models compared to the Crack dataset. This discrepancy indicates that mIoU emphasizes overall pixel-level coverage, whereas Sample\_mIoU focuses more on the accuracy of sample-level predictions. Traditional mIoU typically attains superior scores in managing extensive, readily identifiable areas, although it frequently neglects the identification of minor or sparse anomalies. Sample\_mIoU, via sample-level assessment, guarantees that the categorization decision of each sample equally influences model training. Using KolektorSDD2 as a case study, Sample\_mIoU increased from 60.37\% to 62.08\%. Although it is lower than the mIoU's 61.74\%, this sample-level metric more efficiently reduces bias towards large targets and evaluates detection ability for tiny errors with greater accuracy. Simultaneously, Seg\_Accuracy and Seg\_Recall had occasional declines yet remained comparable or even exhibited overall improvement. Furthermore, Figures \ref{Figure.3}-\ref{Figure.4} depict the visualization outcomes. The data indicate that segmentation outcomes driven by the classification branch more effectively differentiate fine cracks from the background, while traditional fully supervised techniques, such as U-Net, demonstrate somewhat imprecise detail localization. This further illustrates that the classification branch improves segmentation performance by supplying sample-level detection signals, allowing the segmentation network to concentrate on high-risk regions.
\begin{figure*}[htbp]
	\centering
	\includegraphics[width=0.8\textwidth, height=0.55\textheight]{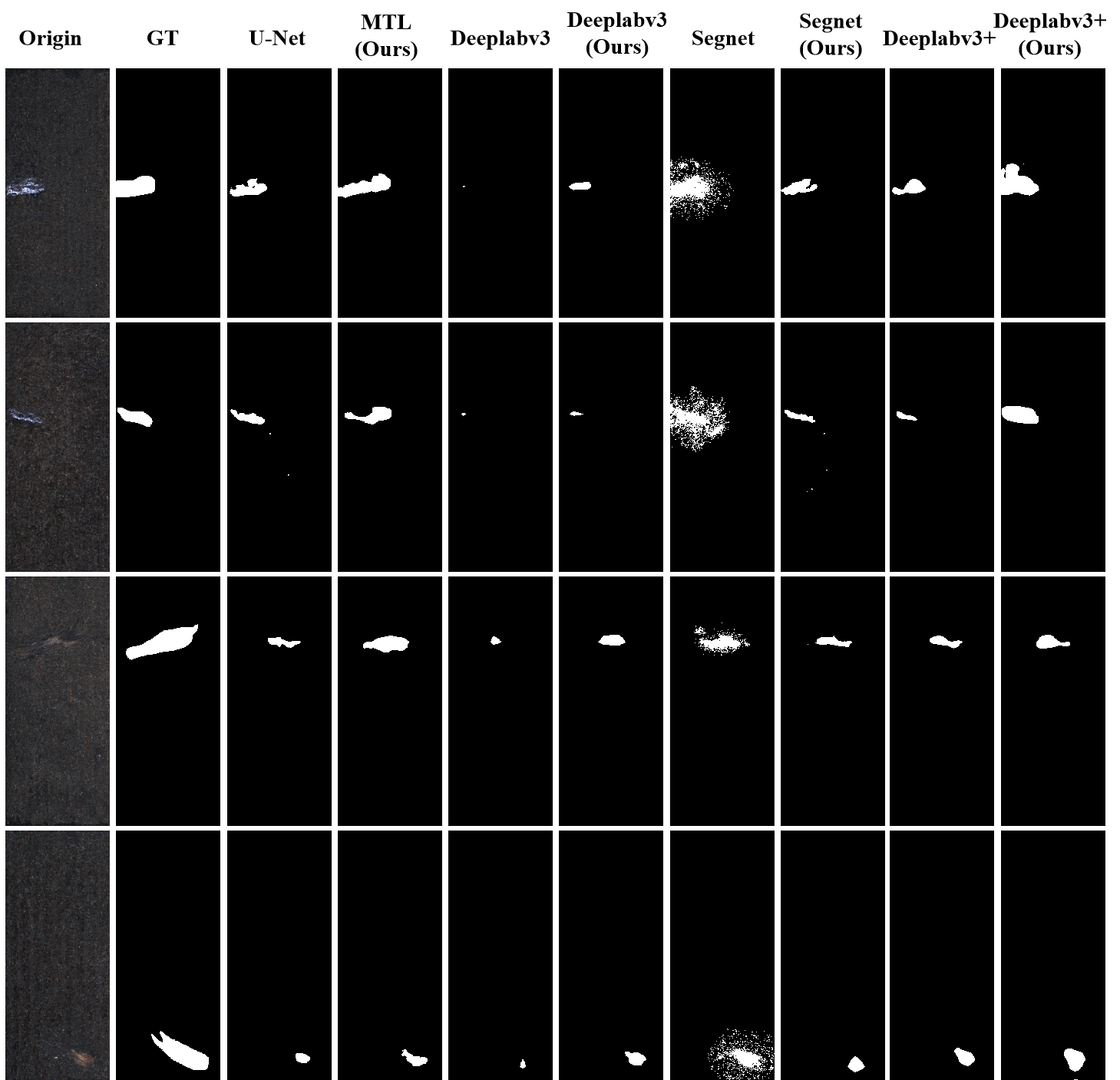} 
	\captionsetup{justification=centering}
	\caption{Ablation Results Visualization of the CaP Plugin on the KSDD2 Dataset}
	\label{Figure.6} 
\end{figure*}

\begin{figure*}[htbp]
	\centering
	\includegraphics[width=0.95\textwidth, height=0.3\textheight]{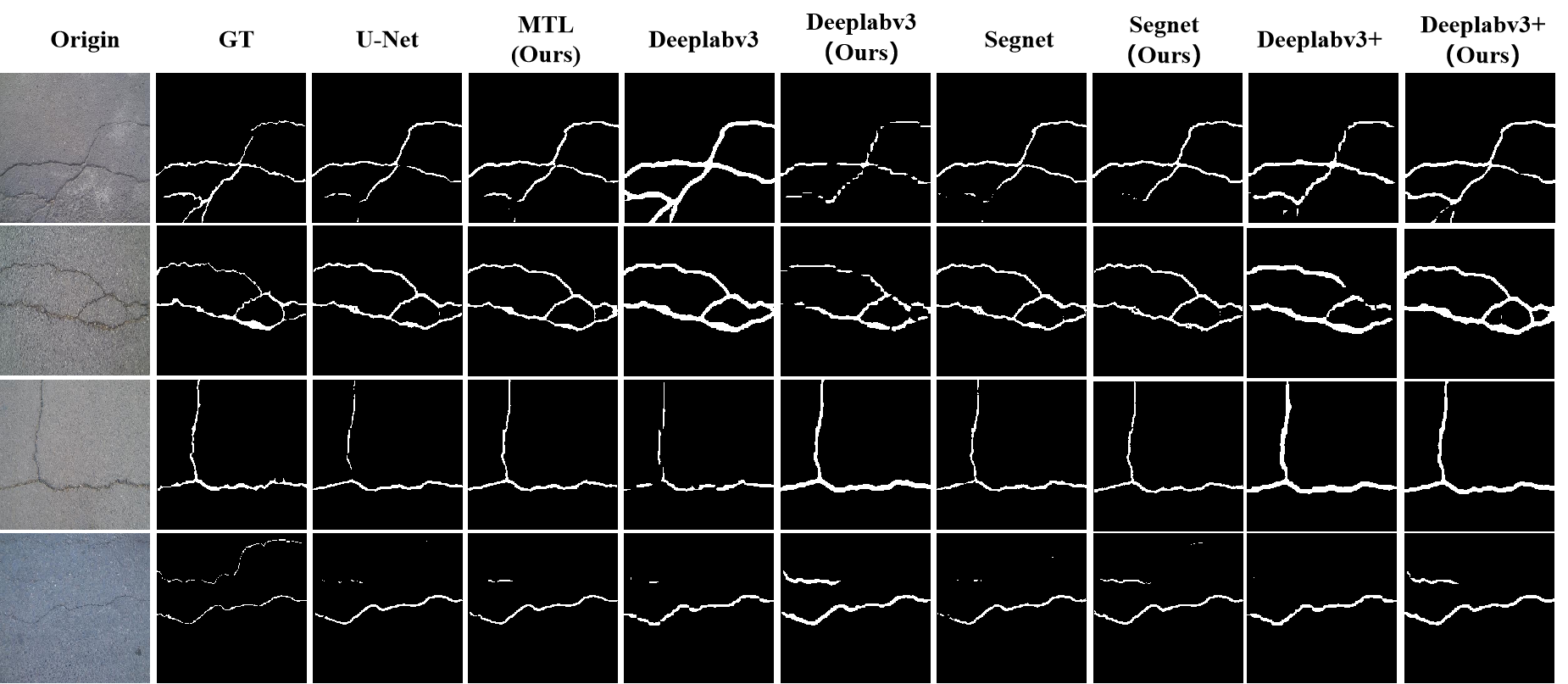} 
	\captionsetup{justification=centering}
	\caption{Ablation Results Visualization of the CaP Plugin on the Crack Dataset}
	\label{Figure.7} 
\end{figure*}
\subsection{Ablation Study}

To assess the universality and steering effect of the CaP plugin across architectures, we conduct a cross-architecture ablation: on representative fully supervised segmentation models (U-Net, DeepLab v3+, and SegNet), we keep the backbone and decoder fixed and strictly align the data split, data augmentation, training epochs, and optimizer. We only attach a lightweight classification branch to the shared encoder and train end-to-end with a joint loss composed of pixel-wise and sample-level (image-level) binary cross-entropy, weighted 0.5 each. This setup controls all variables except “whether the classification branch is present,” allowing performance changes to be attributed to the sample-level supervision and its backpropagated top-down guidance introduced by the classification branch (see Table \ref{tab:compare-ksdd2-crack}). Under this controlled protocol, KSDD2 exhibit more consistent gains in detecting small, low-contrast defects, reflected by improvements in Sample\_mIoU and Seg\_Recall; on the Crack dataset, we observe a slight decrease in Seg\_Accuracy alongside overall improvements in the primary metrics, indicating that in scenarios dominated by large connected components, the sample-level guidance increases sensitivity to incipient crack cues—trading a small rise in false positives for substantive per-item localization and recall gains, which aligns with the risk profile of industrial inspection where missed defects are penalized more heavily (see Figures \ref{Figure.6}–\ref{Figure.7}).

\section{Conclusion}

This work proposes a sample-centric multi-task learning framework and evaluation protocol that effectively mitigates the “small-defect miss” problem in industrial defect inspection caused by the pixel-dominant bias of conventional semantic segmentation. Through sample-level guidance and an evaluation design aligned with per-item quality control, the model achieves higher per-item localization and recall across diverse test scenarios while keeping pixel-level overlap essentially unchanged. For large connected targets (e.g., cracks), the method introduces a modest increase in false positives, leading to a slight drop in Seg\_Accuracy; however, Sample\_mIoU and Seg\_Recall still improve overall—consistent with the engineering trade-off where missed detections are more costly. Future work will focus on adaptive weight scheduling for joint optimization, dynamically balancing the contributions of the segmentation and classification branches based on uncertainty and gradient magnitudes, and incorporating consistency constraints from segmentation to sample-level decisions to reduce threshold sensitivity and false-positive variance. The goal is to further improve Sample\_mIoU and Seg\_Recall without sacrificing pixel-level overlap, strengthening risk alignment with per-item decision-making from an optimization perspective.

\section*{CRediT authorship contribution statement}
\textbf{Hang-Cheng Dong}: Writing – original draft, Visualization, Software, Methodology, Conceptualization, Formal analysis, Investigation, Writing – review \& editing. 
\textbf{Yibo Jiao}: Writing – original draft, Visualization, Software, Formal analysis, Validation, Data curation, Investigation. 
\textbf{Fupeng Wei}: Writing – review \& editing, Validation, Data curation, Investigation. 
\textbf{Guodong Liu}: Supervision, Conceptualization, Formal analysis, Project administration, Resources. 
\textbf{Dong Ye}: Writing – review \& editing, Validation, Data curation, Investigation, Resources. 
\textbf{Bingguo Liu}: Writing – review \& editing, Supervision, Conceptualization, Methodology, Project administration, Resources.

\section*{Declaration of competing interest}
The authors declare that they have no known competing financial interests or personal relationships that could have appeared toinfluence the work reported in this paper.

\section*{Acknowledgements}
This research did not receive any specific grant from funding agencies in the public, commercial, or not-for-profit sectors.

\section*{Data availability}
Data will be made available on request.

\bibliographystyle{elsarticle-num} 
\bibliography{references}






\end{document}